\RequirePackage{tikz}
\documentclass[default]{sn-jnl}

\usepackage{xcolor}

\definecolor{AUC}{HTML}{D7191C}
\definecolor{MRR}{HTML}{FDAE61}
\definecolor{NDCG@5}{HTML}{ABDDA4}
\definecolor{NDCG@10}{HTML}{2B83BA}
\usepackage{array,multirow}
\usepackage{graphicx}
\usepackage[T1]{fontenc}
\usepackage{multirow}
\usepackage{subfigure}
\usepackage{adjustbox}
\usepackage{pgfplots}
\usepackage{graphicx}
\usepackage[labelsep=endash]{caption}
\usepackage{amssymb}
\usepackage{float}
\usepackage{graphicx,stackengine,scalerel}

\usepackage{amssymb}
\usepackage{mathtools}
\usepackage{bm}

\usepackage{pifont}
\newcommand{\xmark}{\ding{55}}%

\jyear{2023}
\begin{document}
\title{Soft Prompt Guided Joint Learning for Cross-Domain Sentiment Analysis}

	\author[1]{\fnm{Jingli} \sur{Shi}}\email{jingli.shi@aut.ac.nz}
	
	\author[1]{\fnm{Weihua} \sur{Li}}\email{weihua.li@aut.ac.nz}

	\author[2]{\fnm{Quan} \sur{Bai}}\email{quan.bai@utas.edu.au}
	
	\author[3]{\fnm{Yi} \sur{Yang}}\email{yyang@hfut.edu.cn}

	\author[4]{\fnm{Jianhua} \sur{Jiang}}\email{jjh@jlufe.edu.cn}

	\affil[1]{\orgdiv{School of Engineering, Computer \& Mathematical Sciences}, \orgname{Auckland University of Technology}, \orgaddress{\street{55 Wellesley Street East}, \city{Auckland}, \postcode{1010}, \country{New Zealand}}}

	\affil[2]{\orgdiv{School of Technology, Environments and Design}, \orgname{University of Tasmania}, \orgaddress{\street{Churchill Ave}, \city{Hobart}, \postcode{7005}, \state{Tasmania}, \country{Australia}}}
	
	\affil[3]{\orgdiv{School of Computer Science}, \orgname{Hefei University of Technology}, \orgaddress{\street{193 Tun Xi Lu, Baohe District}, \city{Hefei}, \postcode{230002}, \state{Anhui}, \country{China}}}

	\affil[4]{\orgdiv{School of Management Science and Information Engineering}, \orgname{Jilin University of Finance and Economics}, \orgaddress{\street{Nanguan District}, \city{Changchun}, \postcode{130117}, \state{Jilin}, \country{China}}}

\abstract{
 Aspect term extraction is a fundamental task in fine-grained sentiment analysis, which aims at detecting customer's opinion targets from reviews on product or service. The traditional supervised models can achieve promising results with annotated datasets, however, the performance dramatically decreases when they are applied to the task of cross-domain aspect term extraction. Existing cross-domain transfer learning methods either directly inject linguistic features into Language models, making it difficult to transfer linguistic knowledge to target domain, or rely on the fixed predefined prompts, which is time-consuming to construct the prompts over all potential aspect term spans. To resolve the limitations, we propose a soft prompt-based joint learning method for cross domain aspect term extraction in this paper. Specifically, by incorporating external linguistic features, the proposed method learn domain-invariant representations between source and target domains via multiple objectives, which bridges the gap between domains with varied distributions of aspect terms. Further, the proposed method interpolates a set of transferable soft prompts consisted of multiple learnable vectors that are beneficial to detect aspect terms in target domain. Extensive experiments are conducted on the benchmark datasets and the experimental results demonstrate the effectiveness of the proposed method for cross-domain aspect terms extraction. 
}


\maketitle

\section{Introduction}
\label{sec:introduction}
\begin{figure}
	\centering
	\includegraphics[width=1.0\textwidth]{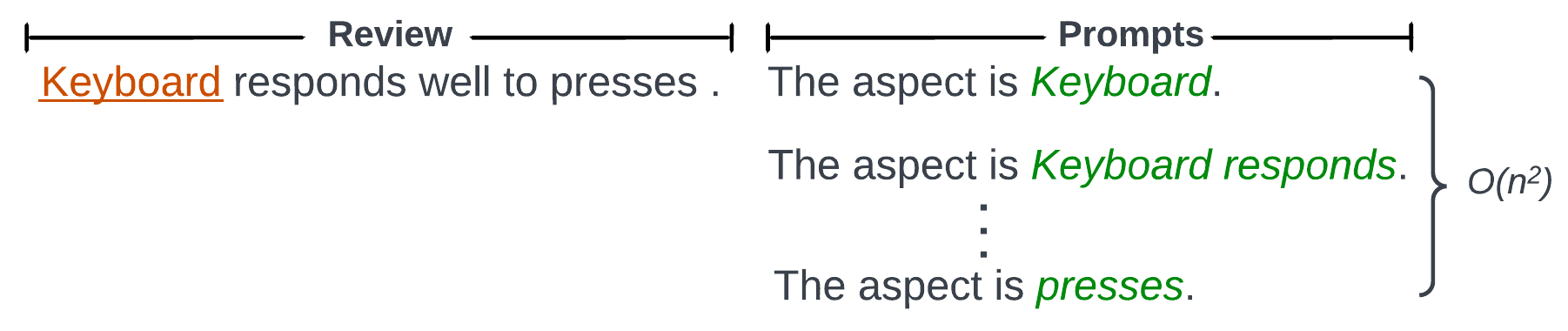}
	\caption{The traditional inputs of prompt tuning model for aspect term extraction. }
	\label{fig:prompt_tuning}   
\end{figure}

To develop specialist knowledge for business development, it's crucial to rapidly understand customer complaints or requirements by analysing their feedbacks. As an outstanding method of review analysis, aspect-based sentiment analysis (ABSA) aims to extract the aspect and opinion terms, and identify their corresponding sentiments from customer reviews\cite{liu2012sentiment, pontiki2016semeval}. In this paper, we focus on  a crucial sub-task for ABSA, named aspect term extraction (ATE), which is to identify opinion targets from customer review sentences. For the example in Figure \ref{fig:prompt_tuning}, the task is expected to detect aspect term \textit{Keyboard} from the sentence ``\textit{Keyboard responds well to presses}".

Recently, ATE has been well studied in literature with the emergence of pre-trained language models (PLMs), such as BERT\cite{devlin2018bert}, BART\cite{lewis2020bart}, T5\cite{raffel2020exploring}, GPT v1-3\cite{radford2018improving, radford2019language, brown2020language}. By fine-tuning PLMs, remarkable results are achieved for ATE task \cite{xu2018double, wang2020combining, wan2020target, gao2021question, venugopalan2022enhanced}. However, a large number of annotated data is required to fine-tune PLMs for downstream tasks. The data annotation work is labour-intensive and time-consuming, which can lead to the lack of training data for fine-tuning PLMs\cite{le2021many}. Moreover, fine-tuning PLMs has become more and more difficulty for real-world applications due to the exponentially increased trainable parameters. To overcome the fine-tuning challenges, a new learning method is designed, named prompt tuning, to reformulate NLP tasks as cloze-style question answering\cite{lester2021power}. Without updating any of the parameters of PLMs, the prompt-based learning has achieved outstanding results on many NLP tasks (e.g., relation classification\cite{chen2022knowprompt}, sentiment classification\cite{li2021sentiprompt}, NER\cite{chen2021lightner}). By manually designing prompts, similar attempts have been made on aspect term extraction to detect aspect term from each text span in a review\cite{gao2022lego, li2022pts, li2021sentiprompt}. As shown in Figure \ref{fig:prompt_tuning}, to extract aspect term \textit{Keyboard}, some prompts are designed using the template \textit{The aspect is \_\_}. Despite the success of fine-tuning and prompt-tuning methods, both of them suffer from the domain challenges. For PLMs in fine-tuning methods, they are pre-trained on universal datasets without specific domains, which leads to task-agnostic and poor performance for domain adaption\cite{xu2019bert}. For prompt-based learning methods, they suffer from high cost on enumerating all possible spans of aspect terms, and the existing models fail to achieve a robust performance on cross domain datasets due to the varied distributions of aspect terms in different domains and the complexity of constructing prompts.

To address the aforementioned challenges in the task of cross domain aspect term extraction, we propose a joint learning method, which is the first to use soft prompt integrating with transferable linguistic knowledge to solve the domain adaptation problem. The designed soft prompts facilitate pre-trained language models better fit the aspect term distributions in target domain, which reduce the high cost due to enumeration of all possible aspect term spans. Following, the linguistic knowledge learning module is designed to learn the domain-invariant representation across domains. By appending learnable prompts and linguistic knowledge with context features, the prompt method can generate better token representations to further improve the performance of cross domain ATE.

The main contributions of this paper can be summarised as follows:

\begin{itemize}
	\item To the best of our knowledge, it's the first to resolve cross domain ATE task by a soft prompt-based joint learning method.
	\item The learnable prompts are designed on multiple source domains to enable efficient knowledge transfer. 
	\item The experimental results demonstrate that the proposed method can outperform the state-of-the-art fine-tuning and prompt-tuning models.  
	
\end{itemize}

The remainder of the paper is organized as follows. In Section \ref{sec:rw}, related works are reviewed in the cross domain aspect term extraction. Section \ref{sec:pr} formally defines the relevant concepts and formulates the problem. The proposed method is introduced in Section \ref{sec:pm}. The experimental work and results are presented and discussed in Section \ref{sec:ea}. Finally, the conclusions and future work are described in Section \ref{sec:cf}.

\section{Related Work}
\label{sec:rw}

Aspect term extraction is a fine-grained sentiment analysis task, which receives a lot of attention. However, only a few studies attempt to focus on domain adaptation for ATE. Cross domain ATE aims to transfer the learned knowledge from the source domain to the target domain which labelled data is limited for ATE task. Due to the complexity of this task and the scarcity of labelled data in target domains, cross domain ATE has become a challenging task. The existing methods can be grouped as three categories: rule-based model, fine-tuning PLM, and prompt-tuning PLM.

such methods may suffer from error propagation induced by entity span detection, high cost due to enumeration of all possible text spans, and omission of inter-dependencies among token labels in a sentence. 

\subsection{Neural Network-based Model}
Early research works about the cross domain ATE mainly focus on hand-crafted domain-independent features and neural network models \cite{ding2017recurrent}. Jakob et al. formulate the ATE problem as an information extraction task, and propose a Conditional Random Field (CRF) based method for single- and cross-domain ATE \cite{jakob2010extracting}. Chernyshevich designs a CRF-based system, which is trained on a mixture of annotated training data, to detect aspect terms on all domain-specific test datasets\cite{chernyshevich2014cross}. For CRF-based methods, they cannot work well if training datasets are from different domains from the test datasets. To overcome this problem, Ding et al. propose a long short-term memory network (LSTM) based method by utilising the domain-independent syntactic rules\cite{ding2017recurrent}. To bridge the gap between different domains, domain-invariant dependency relations are used as pivot information to reduce domain shift by a novel recursive neural network\cite{wang2018recursive}. In the following research, Wang et al. extend the previous work, in which word representations and syntactic head relations are fed into a conditional domain adversarial network\cite{wang2019syntactically}. In another study, Wang et al. exploit local and global memory interactions of an interactive memory network to capture intra-correlations among aspect or opinion terms themselves, as well as between aspect and opinion terms\cite{wang2019transferable}. The auxiliary task and domain adversarial networks are utilised to align source and target space for cross domain ATE.  Marcacini et al. present a transductive learning method to combine features of labelled aspect terms, unlabelled aspect terms, and linguistic information from both source and target domains\cite{marcacini2018cross}. The proposed method can overcome the issue of model inconsistency for cross domain ATE due to different feature spaces. To reduce the reliance of external linguistic resources, an adversarial learning method is presented to learn an alignment weight for each word by aligning the inferred correlation vectors of aspect and opinion terms\cite{li2019transferable}.  Despite the outstanding performance, neural network-based methods fail to obtain satisfactory quality of domain-invariant features and exploit the significant supervision signals in the target domains, which lead to the low precision results.  

\subsection{Language Model-based Model}
Recent research works found that fine-tuning language models with sophisticated task-specific layers can obtain word sense and geometrical dependency parse relations, which benefit the cross domain ATE task\cite{hewitt2019structural}. Pereg et al. incorporate external linguistic information into the language model with a self-attention mechanism for cross domain ATE\cite{pereg2020syntactically}. The proposed method is able to leverage the intrinsic knowledge of language models with externally introduced syntactic features to bridge the gap between source and target domains. Based on BERT, Gong et al. propose an end-to-end framework integrating feature-based adaptation and instance-based adaptation, which significantly improve the performance of language model for ATE\cite{gong2020unified}. Anand et al. apply evolutionary approach to automatically learn linguistic patterns of aspect words, which mitigate the problem of manual engineering pattern rules\cite{anand2021novel}. Mampilli et al. combine language models with attention mechesnism for ATE, and this method achieves good results in-domain and unseen-domain datasets\cite{mampilli2022cross}. Li et al. propose a new generative cross-domain data augmentation framework, which exploits the annotated data from source domain to generate data in target domain for ATE model training\cite{li2022generative}. To solve the model extensibility and robustness on target domain datasets, Howard et al. introduce a novel method to automatically construct domain-specific knowledge graphs of aspect terms, and inject features from these graphs into language models for ATE in target domains\cite{howard2022cross}. Klein et al. utilise syntactic relations connecting opinion and the related aspect words to transfer learned knowledge from language model\cite{klein2022opinion}. Their analyses and experiments prove that the syntactic relations transfer well across domains. To transfer knowledge of aspect terms and sentiment, Dong et al. propose a syntax-base BERT to capture domain-invariant features.  However, all there language model-based methods rely heavily on annotated resources, the performance of fine-tuning language models may be unstable on a small-scale data. Moreover, most of methods only integrate the linguistic features directly to language models, which cannot achieve word-level adaption for aspect extraction.

\subsection{Prompt-based Model}
To address the learning challenges caused by increasing size of LMs, prompt-based methods are proposed to leverage language prompts and task descriptions as context to make ABSA similar to language modelling. Early studies explore hard templates, which are defined manually for ABSA tasks in single domain. Li et al are first to incorporate prompt-based model for aspect-based sentiment analysis subtasks, in which sentiment knowledge prompts are constructed by integrate features from aspects, opinions, and polarities\cite{li2021sentiprompt}. Gao et al. introduce a unified generative framework to solve different ABSA tasks by controlling the type of task prompts\cite{gao2022lego}. By assembling prompts of simple tasks, their method can transfer learned knowledge to difficult tasks. Li et al. propose a prompt-based teacher-student network to alleviate the problem of over-fitting existing in the basic prompt-based models\cite{li2022pts}. Ben et al. present an example-based prompt learning method, which can be applied to unseen domains multiple tasks, namely rumour detection, multi-genre natural language inference, and aspect prediction\cite{ben2022pada}.  However, the domain knowledge is required to design a prompt manually. Therefore, soft prompts constructed to allow LMs to effectively perform specific tasks, which are several learnable vectors instead of human-interpretable natural language. 

Wu et al. adopt soft prompts instead of fixed predefined templates to learn different representations for different domains, then a novel domain adversarial training mechanism to learn domain-invariant features between source domain and target domain for sentiment classification task\cite{wu2022adversarial}. Asai et al. introduce a multi-task language model tuning method that transfers knowledge across different tasks via the soft prompts\cite{asai2022attentional}. Such model is highly parameter-efficient and achieve promising performance using knowledge from high-resource datasets for sentiment classification and other NLP tasks. The existing hard and soft prompt-based methods either focus on single domain or can be only applied to sentiment classification instead of aspect term extraction.

In this paper, to alleviate the challenges of cross domain aspect term extraction in the existing models, a joint learning method is proposed to integrate high-quality transferable knowledge from source domains via a mixture of trainable soft prompts and domain-invariant and learnable linguistic features. Different from the previous works, the proposed method is the first work that incorporates soft prompts into jointly training to solve the cross domain ATE problem. The soft prompts can overcome the time-consuming issue caused by hard prompts for enumerating the prompt queries over all potential aspect spans. The learnable linguistic features can serve as an enhancement component to bridge the gap between different domains and further capturing domain-invariant features for ATE task. The proposed method enables efficient knowledge transfer from source domains and achieve outstanding performance on multiple datasets for cross domain ATE. Furthermore, the analysis of experimental results shows that the soft prompts and learnable syntactic features largely contribute to the performance improvements.

\section{Preliminaries}
\label{sec:pr}
In this section, the formal definitions related to cross domain ATE are presented, and then the problem is formally formulated based on these definitions. 
\subsection{Problem Formulation}
Formally, the proposed method formulate the task of cross-domain aspect term extraction as a sequence tagging problem. Two domain datasets are given, $\mathbb{D}_s$ and $\mathbb{D}_t$ which represent the source and target domain, respectively. For the source domain dataset, $\mathbb{D}_s = \{S_s^i, y_s^i \}_{i=1}^{N_s}$ are $N_s$ annotated sentences, where $S_s^i$ is the $i$th sentence, $y_s^i \in \{ B, I, O\}$ denotes the corresponding aspect label. In target domain dataset, $\mathbb{D}_t = \{S_t^i  \}_{i=1}^{N_t}$ consists of $N_t$ unlabelled sentences, where $S_t^i$ indicates the $i$th sentence. The goal of cross domain ATE is to learn a function, which can learn both in-domain and domain-invariant knowledge between source and target domain to better predict token-level labels on the test set from the target domain.

\section{Soft Prompt-Based Joint Learning Model} 
\label{sec:pm}
In this section, we first describe the overview of the proposed method. Then we introduce each module from bottom to up in the whole architecture. Finally, we present the learning objective for cross-domain ATE. 

\begin{figure}
	\centering
	\includegraphics[width=1.0\textwidth]{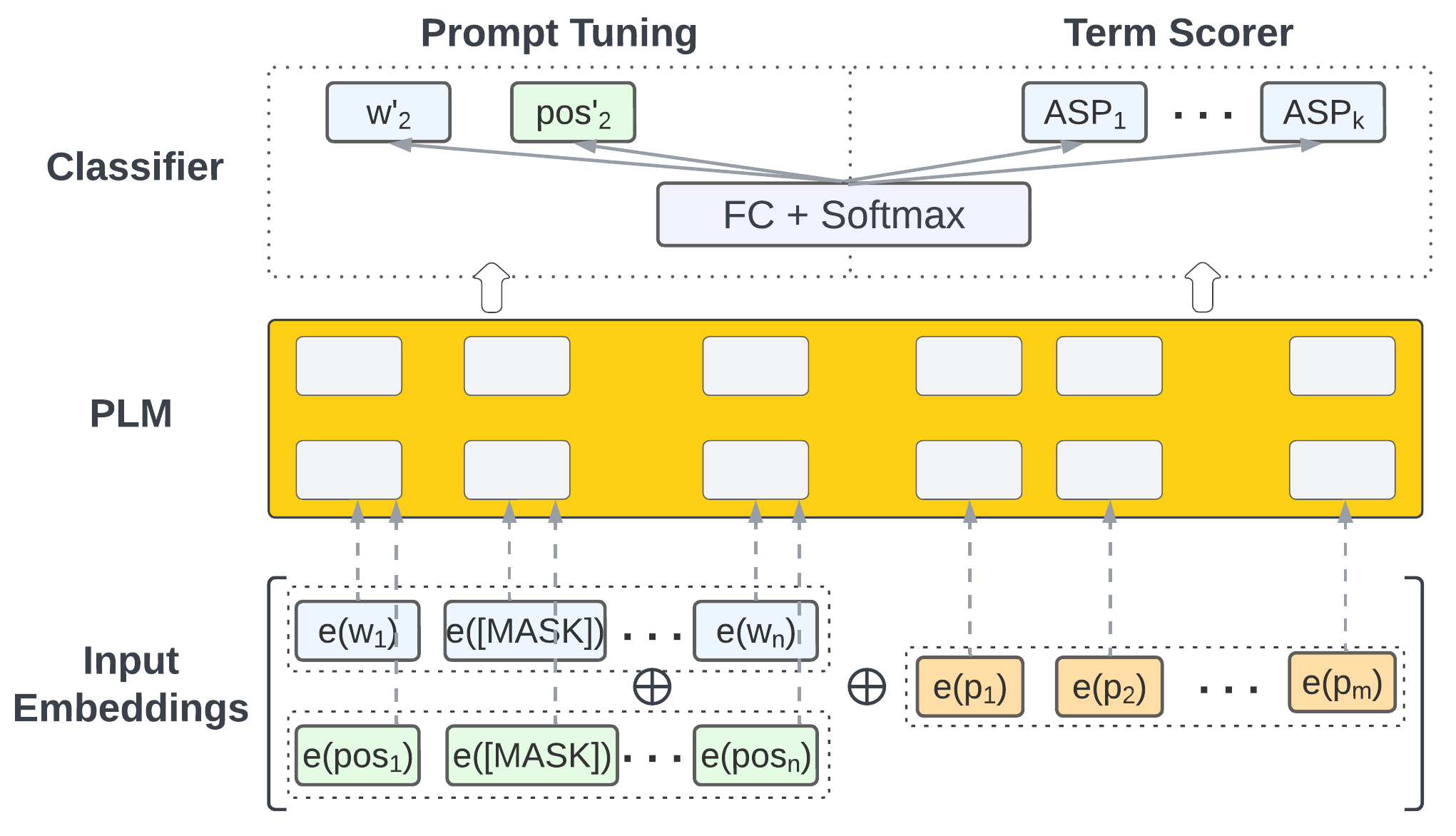}
	\caption{Overview architecture of the soft prompt-based joint learning model.}
	\label{fig:overview}   
\end{figure}

The overall architecture for our feature-based domain adaptation component is shown in Figure \ref{fig:overview}. Together with word and syntax embeddings, generated prompts are encoded as features and are fed into pre-trained language model. The output representations are as input of Softmax layer. Except for the apsect term extractor, a syntax learning module is designed to learn structural correspondence between domains. Each module is described in the following sub-sections. 

\subsection{Input Embedding}
Given a sentence $s = \{w_1, w_2, ..., w_n\}$ with $n$ words, the word sequences are converted into continuous embedding $E_s = \{e_1, e_2, ..., e_n\}$. For each embedding $e_i$, it consists of three type embeddings: (a) word embedding $e^w$ is obtained via pre-trained language model by Equation (\ref{eq:emb_w}). (b) syntax embedding $e^{pos}$ is calculated in Equation (\ref{eq:emb_pos}). To leverage the domain-invariant features more effectively, 25\% of original POS tags are randomly replaced with a special token [MASK], and a syntax learning module is design to predict the masked POS tags. (c) soft prompt embedding are computed in Equation (\ref{eq:emb_p}). Inspired by previous work on prompt features \cite{ziser2018pivot, ben2020perl, ben2021pada}, Mutual Information is applied to automatically extract prompts. To select prompts that are related all source domains, the Euclidean distance is computed on T5 embeddings of prompts and the aspect tokens to generate $m$ features for each training input. 

\begin{equation}
	\label{eq:emb_w}	
	e^w = T5(\{w_1, w_2, ..., w_n\})
\end{equation}	

\begin{equation}
	\label{eq:emb_pos}	
	e^{pos} = T5(\{t_1, [MASK], ..., t_n\})
\end{equation}	

\begin{equation}
	\label{eq:emb_p}	
	e^{p} = T5(\{p_1,  ..., p_m\})
\end{equation}	

\subsection{Soft Prompt Learning}
Prompt tuning is a method integrating extra information into pre-trained language models by converting downstream tasks into cloze questions. The prompt is the primary component for prompt tuning model. In the proposed method, prompts are aspect terms encoding domain-specific semantics. We leverage the prompts from various domains to span the shared semantic space, and reflect the similarities and differences between different domains. 

The prediction of aspect term is formalised with designed prompts in Equation (\ref{eq:y_p}). 

\begin{equation}
	\label{eq:y_p}	
	\hat{y}^{p} = softmax(W^{p}*[e^w ; e^{pos} ; e^{p}] + b^{p}),
\end{equation}	
where $W^{p}$ is the training weights and $b^{p}$ is the bias vector. The training objective of soft prompt tuning is calculated using cross-entropy loss in Equation (\ref{eq:los_p}).

\begin{equation}
	\label{eq:y_p}	
	\hat{y}^{p} = softmax(W^{p}*[e^w ; e^{pos} ; e^{p}] + b^{p}),
\end{equation}	

\begin{equation}
	\label{eq:los_p}	
	\mathscr{L}_{prompt} =  \sum^{\mathbb{D}_s}\sum_{i}^{n}f(\hat{y}^{p}_i, y^{p}_i),
\end{equation}	

\subsection{Syntax Learning}
\begin{figure}
	\centering
	\includegraphics[width=1.0\textwidth]{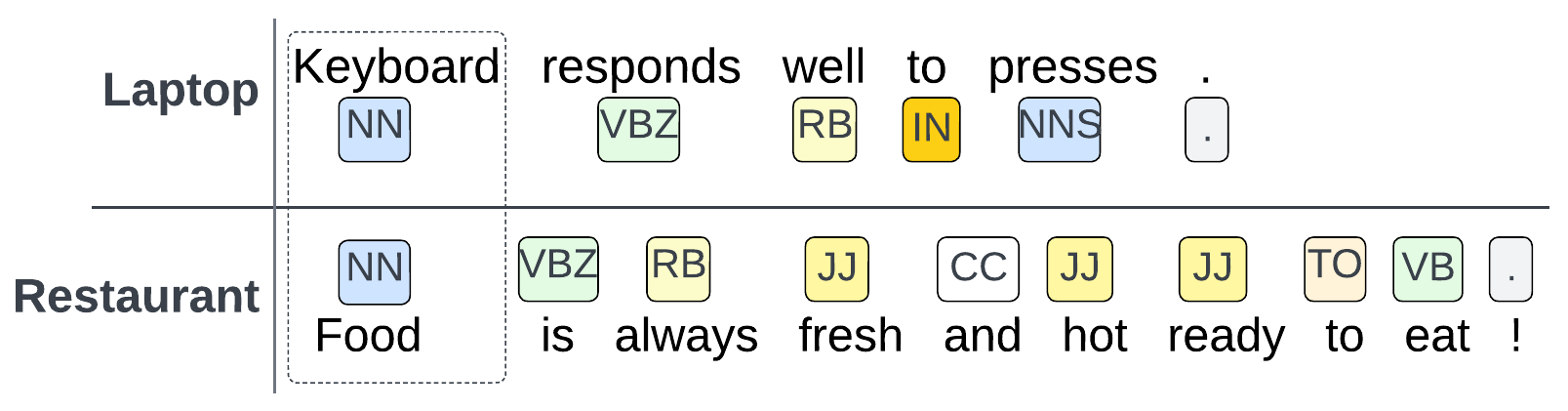}
	\caption{POS tags of reviews from laptop and restaurant domain. The aspect terms \textit {Keyboard} and \textit {Food} share same POS tag \textit {NN}. }
	\label{fig:pos}   
\end{figure}

For aspect terms from different domains, their linguistic features maintain often-occurring pattern\cite{hu2004mining, qiu2011opinion, chen2021bridge}. Given an example in Figure \ref{fig:pos}, The aspect \textit{Keyboard} from domain \textbf{Laptop} shares same POS tag \textit{NN} with the aspect \textit{Food} in domain \textbf{Restaurant}, indicating that these aspect terms are similar in syntax. To learn the syntax knowledge, the encoded masked feature $e^{pos}$ is fed into a softmax layer. The predicted POS tag can be calculated in Equation (\ref{eq:y_pos}).

\begin{equation}
	\label{eq:y_pos}	
	\hat{y}^{pos} = softmax(W^{pos}*[e^w ; e^{pos} ; e^{p}] + b^{pos}),
\end{equation}	
where $\hat{y}^{pos} \in \mathbb{R}^{N^{pos}}$, $N^{pos}$ is the number of total POS tags. $W^{pos}$ is the learnable weight, and $b^{pos}$ is the bias tensor. To optimise the learning process, the cross-entropy loss is calculated in Equation (\ref{eq:los_pos}).

\begin{equation}
	\label{eq:los_pos}	
	\mathscr{L}_{syntax} =  \sum^{\mathbb{D}_s}\sum_{i}^{n}I(i)*f(\hat{y}^{pos}_i, y^{pos}_i),
\end{equation}	

\begin{equation}
	\label{eq:los_pos_i}	
	I(i) = \left\{\begin{matrix}
		1 &  if \, token \, is \, masked \\ 
		0 & else
	\end{matrix}\right. 
\end{equation}	
where $I(i)$ is the indicator to filter the masked tokens. $y^{pos}_i$ is the real POS tag of $i$th token in the input sentence.

\subsection{Training Objective}
Given the source domain datasets and the target dataset, the aspect term extraction and syntax discriminator are jointly trained for optimising the soft prompt embeddings, syntax embedding, and aspect term predictor. The final training objective is obtained by weighted sum of the cross-entropy losses from syntax learning and multiple domain knowledge enhanced prompt-tuning in Equation (\ref{eq:loss_final}).

\begin{equation}
	\label{eq:loss_final}	
	\mathscr{L}(\theta) = \alpha * \mathscr{L}_{prompt} + \beta  * \mathscr{L}_{syntax}
\end{equation}	
where $\alpha$ and $\beta$ are the trade-off parameter.

\section{Experiments}
\label{sec:ea}
In this section, extensive experiments are conducted on three group of datasets to evaluate the proposed model

\subsection{Dataset} 
The experiments of our method are conducted on two group of benchmark datasets with different domains. For first group of dataset $\mathbb{G}$1, four domains are included: Device ($\mathbb{D}$) is the set of all the digital device reviews\cite{toprak2010sentence}. Laptop ($\mathbb{L}$) and Restaurant ($\mathbb{R}$) are from SemEval ABSA challenges\cite{pontiki2014semeval, pontiki2015semeval, pontiki2016semeval}, which contain customer reviews of laptop and restaurant. Service ($\mathbb{S}$) refer to the customer reviews of web services\cite{hu2004mining}. The basic statistics of the first group of dataset are presented in Table \ref{tb:stat_d1}. The second group of datasets $\mathbb{G}$2 contains three domains: Diapers($\mathbb{DI}$), Antivirus Software ($\mathbb{AS}$) and Electronics($\mathbb{E}$), shown in Table \ref{tb:stat_d2}. $\mathbb{DI}$ and $\mathbb{AS}$ are prepared for opinion mining\footnote{https://www.cs.uic.edu/~liub/FBS/Reviews-9-products.rar} by\cite{ding2008holistic}. $\mathbb{E}$ is annotated by our annotator, which is originally collected for cross-domain sentiment classification\footnote{https://github.com/paolazola/Cross-source-cross-domain-sentiment-analysis} by \cite{zola2019social}.

\begin{table}
	\centering
	\caption{Statistics of the first group datasets.}
	\label{tb:stat_d1}
	\begin{tabular}{l|ccc} 
		\hline
		\textbf{Domain} & \begin{tabular}[c]{@{}c@{}}\textbf{Sentences}\\\end{tabular} & \textbf{Train} & \textbf{Test}  \\ 
		\hline
		$\mathbb{R}$    & 6035                                                         & 3877           & 2158           \\
		$\mathbb{L}$    & 3845                                                         & 3045           & 800            \\
		$\mathbb{D}$    & 3836                                                         & 2557           & 1279           \\
		$\mathbb{S}$    & 2239                                                         & 1492           & 747            \\
		\hline
	\end{tabular}
\end{table}

\begin{table}
	\centering
	\caption{Statistics of the second group datasets.}
	\label{tb:stat_d2}
	\begin{tabular}{l|ccc} 
		\hline
		\textbf{Domain} & \begin{tabular}[c]{@{}c@{}}\textbf{Sentences}\\\end{tabular} & \textbf{Train} & \textbf{Test}  \\ 
		\hline
		$\mathbb{DI}$   & 375                                                          & 262            & 113            \\
		$\mathbb{AS}$   & 380                                                          & 266            & 114            \\
		$\mathbb{E}$    & 550                                                          & 385            & 165    		  \\  
		\hline  
	\end{tabular}
\end{table}

\subsubsection{Implementation and Hyper-parameters}
In the proposed method, the Pytorch framework \footnote{https://pytorch.org/} is utilised to implement our model. T5-base \footnote{https://huggingface.co/t5-base} is used as the base LMs. We use Stanford NLP Toolkit \footnote{https://stanfordnlp.github.io/stanza/}(i.e., Stanza \cite{qi2020stanza}) to obtain the syntax structures of all datasets. All experiments are conducted on a single NVIDIA RTX A6000 GPU accelerator.

The default settings are used for T5-base, e.g., 24 layers of self-attention with 1024 dimensional hidden vectors. The Adam optimiser \cite{kingma2015adam} is applied with an initial learning rate of 2e-3. The epoch is set to 20, and the batch size is 16.

\subsection{Baselines}  
To verify the effectiveness of the proposed method, several competitive baselines are utlised to compare with our model.  

\begin{itemize}
	\item \textbf{\emph{CrossCRF}}\cite{jakob2010extracting} is a traditional sequence labelling method, which linguistic features (i.e., word type, POS tag and dependency relation) are applied to detect aspect terms using CRF.
	\item \textbf{\emph{DP}}\cite{qiu2011opinion} addresses two problems, i.e., opinion lexicon expansion and opinion target extraction using a semi-supervised method based on bootstrapping. The dependency relations linking opinion terms and targets are extracted using a dependency parser, and then the identified relations are used to expand the initial opinion lexicons and detect aspect terms.
	\item \textbf{\emph{mSDA}}\cite{chen2012marginalized} is a marginalised stacked denoising auto-encoder, which uses linear denoisers to build blocks for learning feature representations. This method can address the issues of high computational  cost and lack of scalability to high-dimensional features.    
	\item \textbf{\emph{FEMA}}\cite{yang2015unsupervised} performs dense feature representations learning, which are more robust to domain shift, using neural language models to obtain low-dimensional embeddings directly. 
	\item \textbf{\emph{RNCRF}}\cite{wang2016recursive} integrates recursive neural networks and CRFs into a joint model to detect aspect and opinion terms. The unified framework can propagate bidirectional information between aspect and opinion terms, and learn high-level discriminative features. 
	\item \textbf{\emph{Hier-Joint}}\cite{ding2017recurrent} combines rule-based, unsupervised aspect term extraction with neural network based supervised methods to learn a hidden representation for different domains. 
	\item \textbf{\emph{RNSCN}}\cite{wang2018recursive} is a novel recursive neural network, which can reduce the issue of domain shift in word level by dependency relations. The syntactic relations can be used as invariant pivot information across different domains between source and target datasets.  
	\item \textbf{\emph{AD-SAL}}\cite{li2019transferable} firstly explores an unsupervised domain adaption setting for joint extraction of aspect and opinion terms. Moreover, a selective adversarial learning method is proposed to learn an alignment weight for each word to achieve fine-grained domain adaption.  
	\item \textbf{\emph{BERT}} directly fine-tunes base BERT\cite{devlin2018bert} to predict collapsed labels for cross-domain ATE task.  
	\item \textbf{\emph{TRNN-GRU}}\cite{wang2019syntactically} introduces a conditional domain adversarial network to improve the knowledge transferability across different domains. Furthermore, the recursive neural network with a sequence labelling classifier is integrated to model contextual influence to predict the aspect terms in target datasets. 
	\item \textbf{\emph{CrossBERT}}\cite{xu2019bert} post-trains base BERT\cite{devlin2018bert} on mixed datasets from Yelp and Amazon reviews, and then fine-tunes the trained model to detect aspect terms across domains. 
	\item \textbf{\emph{CrossBERT-UDA}}\cite{gong2020unified} is an end-to-end framework that performs feature and instance based adaption for cross-domain ABSA tasks. This method can learn domain-invariant features via linguistic information, and perform word-level instance weighting based on BERT.
	\item \textbf{\emph{SA-EXAL}}\cite{pereg2020syntactically} incorporates external linguistic information into a self-attention mechanism with BERT, which can bridge the gap across domains by leveraging the intrinsic knowledge from BERT  with external syntactic information.
	\item \textbf{\emph{CDRG-Indep}}\cite{yu2021cross} aims to generate target-domain data with fine-grained annotation based on labelled data in source domain, and then directly train a sequence labelling model on the generated dataset by adopting BERT model. 
	\item \textbf{\emph{CDRG-Merge}}\cite{yu2021cross} is similar with CDRG-Indep except for the training strategy, which merges the labelled source data with generated data as training examples.
	\item \textbf{\emph{AHF}}\cite{zhou2021adaptive} integrates pseudo-label based semi-supervised learning and adversarial training in a unified network for cross domain ABSA tasks. The target data is utilised for training domain discriminator and refine the task classifier.
	\item \textbf{\emph{SynBridge}}\cite{chen2021bridge} is an active domain adaptation model that transfers aspect words by actively supplementing transferable knowledge. The syntactic bridges are constructed via recognising syntactic roles as pivots to identify transferable syntactic roles for the words across domains. 
	\item \textbf{\emph{SemBridge}}\cite{chen2021bridge} is similar model with SynBridge, but SemBridge retrieves transferable prototypes to link aspect words across domains. 
	\item \textbf{\emph{SDAM}}\cite{dong2022syntax} is a syntax-guided domain adaptation method that exploits syntactic structure similarities to build pseudo training data. 
	\item \textbf{\emph{FMIM-BERT}}\cite{chen2022simple} is a simple but effective method based on mutual information maximization for cross-domain ABSA tasks. 
\end{itemize}	

\subsection{Experimental Results and Model Analysis}
We conduct experiments of cross-domain ATE on two group of datasets $\mathbb{G}$1 and $\mathbb{G}$2, and the overall comparison results are shown in Table \ref{table:baselines_g1} and \ref{table:baselines_g2}. We can observe that our method outperforms all baselines on most domains in dataset $\mathbb{G}$1 and all domains in dataset $\mathbb{G}$2. Compared with the previous approaches, our method is significantly superior to machine learning and based models. However, the performance of our method is lower than that of SemBridge for domain adaption $\mathbb{L}$ $\to$ $\mathbb{R}$ and $\mathbb{D}$ $\to$ $\mathbb{R}$ (-0.007 and -0.006, respectively), indicating that SemBridge captures more syntactic and semantic knowledge of source domains and transfers these meaningful knowledge to target domain. Without semantic features, SynBridge achieves a degraded performance compared with the proposed method. Whereas, our method can outperform most of fine-tuning based models, which shows that soft prompt tuning based method can learn domain-dependent features, but also domain-invariant knowledge. The outstanding performance demonstrates the prompt-tuning based method is able to solve the problem of cross-domain ATE.  

\begin{table}
	\centering
	\caption{Experimental results for cross-domain ATE on $\mathbb{G}$1}
	\label{table:baselines_g1}
	\resizebox{\textwidth}{!}{
		\begin{tabular}{l|ccc|ccc|ccc|ccc} 
			\hline
			\textbf{Model}         & $\mathbb{R}$ $\to$ $\mathbb{L}$   & $\mathbb{S}$ $\to$ $\mathbb{L}$   & $\mathbb{D}$ $\to$ $\mathbb{L}$   & $\mathbb{L}$ $\to$ $\mathbb{R}$   & $\mathbb{S}$ $\to$ $\mathbb{R}$   & $\mathbb{D}$ $\to$ $\mathbb{R}$   & $\mathbb{R}$ $\to$ $\mathbb{D}$   & $\mathbb{L}$ $\to$ $\mathbb{D}$   & $\mathbb{S}$ $\to$ $\mathbb{D}$   & $\mathbb{R}$ $\to$ $\mathbb{S}$   & $\mathbb{L}$ $\to$ $\mathbb{S}$   & $\mathbb{D}$ $\to$ $\mathbb{S}$    \\ 
			\hline\hline
			CrossCRF      & 0.197 & 0.116 & 0.242 & 0.282 & 0.170 & 0.659 & 0.211 & 0.299 & 0.097 & 0.088 & 0.086 & 0.045  \\
			DP            & 0.198 & 0.198 & -     & 0.376 & 0.376 & 0.376 & 0.218 & -     & 0.218 & 0.197 & 0.197 & 0.197  \\
			mDA           & 0.209 & 0.146 & 0.257 & 0.243 & 0.325 & 0.213 & 0.172 & 0.294 & 0.169 & 0.131 & 0.131 & 0.131  \\
			FEMA          & 0.266 & 0.150 & 0.268 & 0.350 & 0.376 & 0.207 & 0.229 & 0.296 & 0.187 & 0.108 & 0.148 & 0.088  \\
			RNCRF         & 0.243 & -     & 0.406 & 0.409 & -     & 0.346 & 0.243 & 0.315 & -     & -     & -     & -      \\ 
			\hline
			Hier-Joint    & 0.317 & 0.300 & 0.362 & 0.467 & 0.520 & 0.504 & 0.320 & 0.316 & 0.334 & 0.198 & 0.234 & 0.235  \\
			RNSCN         & 0.266 & 0.189 & -     & 0.356 & 0.332 & 0.346 & 0.333 & -     & 0.220 & 0.200 & 0.166 & 0.200  \\
			AD-SAL        & 0.341 & 0.270 & -     & 0.430 & 0.410 & 0.410 & 0.354 & -     & 0.336 & 0.280 & 0.272 & 0.266  \\
			BERT          & 0.314 & 0.305 & -     & 0.404 & 0.447 & 0.403 & 0.276 & -     & 0.339 & 0.195 & 0.258 & 0.303  \\
			TRNN-GRU      & 0.402 & -     & 0.517 & 0.538 & -     & 0.512 & 0.373 & 0.412 & -     & -     & -     & -      \\
			CrossBERT     & 0.397 & 0.350 & -     & 0.454 & 0.513 & 0.426 & 0.332 & -     & 0.332 & 0.244 & 0.233 & 0.282  \\
			CrossBERT-UDA & 0.439 & 0.348 & -     & 0.495 & 0.471 & 0.427 & 0.349 & -     & 0.321 & 0.331 & 0.279 & 0.280  \\
			SA-EXAL       & 0.476 & -     & 0.477 & 0.547 & -     & 0.545 & 0.405 & 0.422 & -     & -     & -     & -      \\
			CDRG-Indep    & 0.402 & 0.332 & -     & 0.551 & 0.538 & 0.501 & 0.308 & -     & 0.349 & 0.417 & 0.441 & 0.371  \\
			CDRG-Merge    & 0.466 & 0.395 & -     & 0.600 & 0.563 & 0.527 & 0.326 & -     & 0.369 & 0.424 & 0.471 & 0.418  \\
			AHF           & 0.557 & 0.448 & -     & 0.646 & 0.591 & 0.597 & 0.502 & -     & 0.478 & 0.438 & 0.427 & 0.444  \\
			SynBridge     & 0.551 & 0.453 & -     & 0.653 & 0.584 & 0.628 & 0.533 & -     & 0.539 & 0.327 & 0.337 & 0.381  \\
			SemBridge     & 0.579 & 0.451 & -     & 0.662 & 0.593 & 0.636 & 0.553 & -     & 0.546 & 0.350 & 0.350 & 0.377  \\
			SDAM          & 0.546 & 0.467 & -     & 0.631 & 0.586 & 0.609 & 0.516 & -     & 0.580 & 0.456 & 0.453 & 0.552  \\
			FMIM-BERT     & 0.494 & 0.424 & -     & 0.634 & 0.592 & 0.573 & 0.397 & -     & 0.376 & 0.514 & 0.549 & 0.528  \\
			Ours          & 0.593 & 0.480 & 0.527 & 0.655 & 0.612 & 0.630 & 0.563 & 0.434 & 0.586 & 0.528 & 0.555 & 0.570   \\
			\hline
		\end{tabular}
	}
\end{table}

Table \ref{table:baselines_g2} presents the results of cross-domain ATE on a small scale of dataset. Compared with fine-tuning based models, AD-SAL and CrossBERT-UDA, our method achieves the best performance on all three domains (over 5\% absolute improvement). The improvement demonstrates that the prompt tuning based mode can be applied to both big and small scale of datasets with competitive performance. Compared with prompt tuning models, it's more difficult to train the domain-specific model on source domains for fine tuning approaches. While prompt tuning models can activate some prior knowledge in language models by the feature distribution of prompts.

\begin{table}
	\centering
	\caption{Experimental results for cross-domain ATE on $\mathbb{G}$2.}
	\label{table:baselines_g2}
	\resizebox{\textwidth}{!}{
		\begin{tabular}{l|cc|cc|cc} 
			\hline
			\textbf{Model} & 
			$\mathbb{DI}$ $\to$ $\mathbb{AS}$  & 
			$\mathbb{E}$ $\to$ $\mathbb{AS}$ & 
			$\mathbb{DI}$ $\to$ $\mathbb{E}$ & 
			$\mathbb{AS}$ $\to$ $\mathbb{E}$ & 
			$\mathbb{E}$ $\to$ $\mathbb{DI}$ & 
			$\mathbb{AS}$ $\to$ $\mathbb{DI}$ \\ 
			\hline\hline
			AD-SAL         & 0.208   & 0.181                & 0.172                & 0.191                & 0.156  & 0.176    \\
			CrossBERT-UDA  & 0.244   & 0.201                & 0.195                & 0.213                & 0.160  & 0.196    \\
			Ours           & 0.322   & 0.312                & 0.279                & 0.313                & 0.213  & 0.251    \\
			\hline
		\end{tabular}
	}
\end{table}

\subsection{Further Analysis}
\label{subsec:fa}

\subsection{Ablation Study}
To analyse the effect of each component including syntax learning and prompts, the ablation experiments are conducted on dataset $\mathbb{G}$2 and the experimental results are shown in Table \ref{tb:ablation}.

\begin{table}
	\centering
	\caption{Ablation study over cross-domain ATE on $\mathbb{G}$2. w/o indicates without.}
	\label{tb:ablation}
	\resizebox{\textwidth}{!}{
		\begin{tabular}{l|cc|cc|cc} 
			\hline
			\textbf{Model} & $\mathbb{DI}$ $\to$ $\mathbb{AS}$  & 
			$\mathbb{E}$ $\to$ $\mathbb{AS}$ & 
			$\mathbb{DI}$ $\to$ $\mathbb{E}$ & 
			$\mathbb{AS}$ $\to$ $\mathbb{E}$ & 
			$\mathbb{E}$ $\to$ $\mathbb{DI}$ & 
			$\mathbb{AS}$ $\to$ $\mathbb{DI}$  \\ 
			\hline\hline
			-w/o Syntax    & 0.267   & 0.279                & 0.241                & 0.273                & 0.198  & 0.218    \\
			-w/o Prompts   & 0.244   & 0.258                & 0.261                & 0.253                & 0.171  & 0.226    \\
			only T5        & 0.196   & 0.218                & 0.133                & 0.236                & 0.119  & 0.164    \\
			Ours           & 0.322   & 0.312                & 0.279                & 0.313                & 0.213  & 0.251    \\
			\hline
		\end{tabular}
	}
\end{table}

\subsubsection{Effect of Syntax Learning}
In this subsection, the effect of syntax learning is verified via ablation study. Table \ref{tb:ablation} presents the experimental results on dataset $\mathbb{G}$2. We find that, without syntax learning component, the results of our method see an decrease on all target domains (i.e., -5.5\%, -3.3\%, -3.8\%, -4\%, -1.5\%, -3.3\%, respectively). This shows that the linguistic features are necessary to capture domain-invariant information between domains. The domain independent features can bridge the gap over domains and facilitate the prediction in target domain for cross-domain ATE task. For example, the model trained on domain $\mathbb{DI}$ with one input sentence ``\textit{The Diaper Champ is the best we found!}", and the aspect term is \textit{Diaper Champ} with POS tag \textit{NN}. After learning this syntactic pattern in source domain, it can be easier for the model to predict aspect term \textit{Program} in the sentence ``\textit{The program brings more problems than a virus...}" on target domain $\mathbb{AS}$. 

\subsubsection{Effect of Prompts}
We present an evaluation on the effect of prompts by removing them from our method. In Table \ref{tb:ablation}, the F1 scores of w/o prompts are presented across all target domains. After removing prompts, we observe a significant performance drop on all domains (-7.8\%, -5.4\%, -1.8\%, -6.0\%, -4.2\%, -2.5\%, respectively), suggesting the designed prompts can be leveraged to span the semantic space of source domains. The shared semantic knowledge can further promote the performance for cross-domain ATE task. 

\begin{figure}
	\centering
	\includegraphics[width=0.6\textwidth]{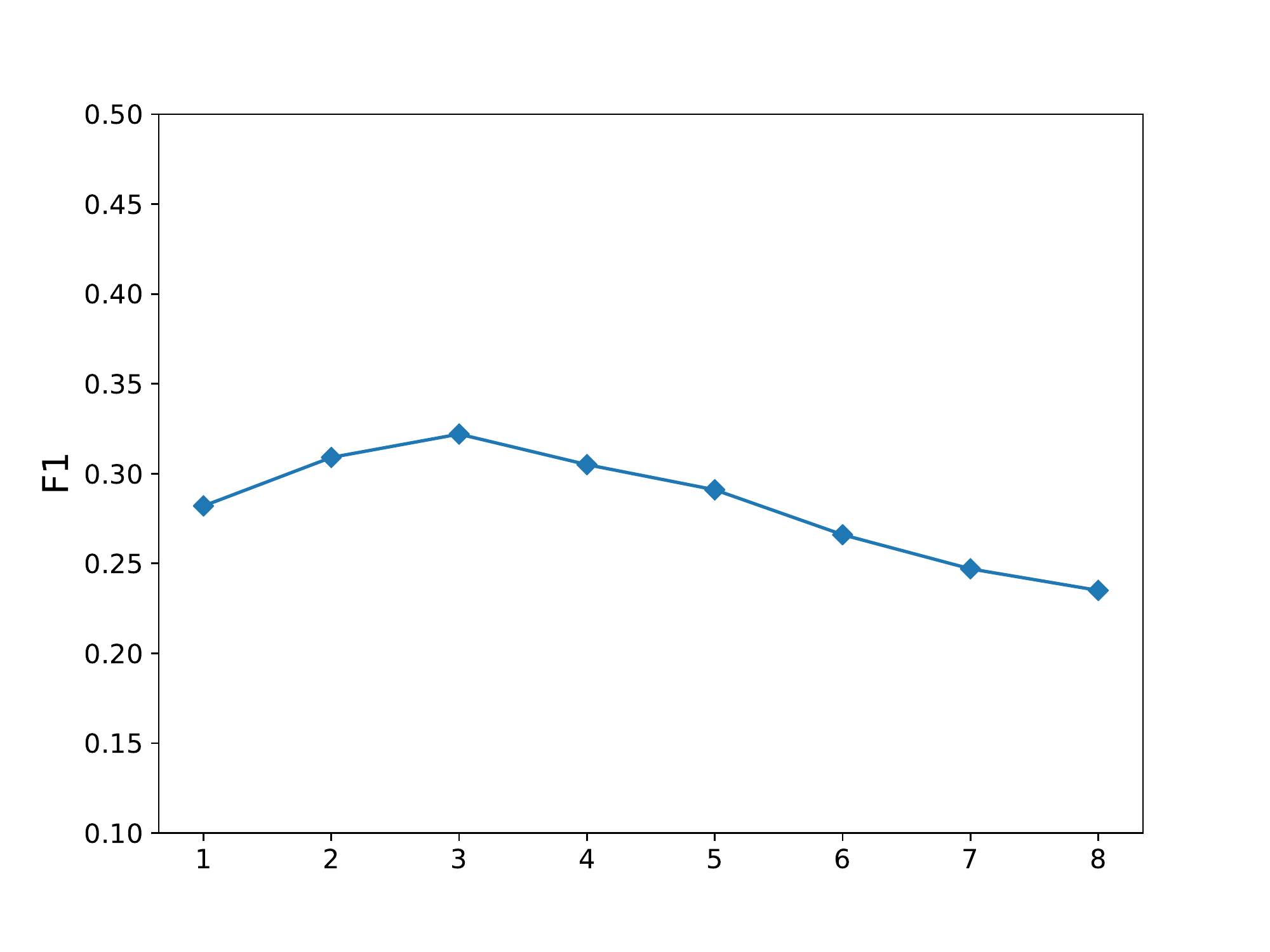}
	\caption{Experimental results of different lengths of soft prompt tokens on $\mathbb{DI}$ $\to$ $\mathbb{AS}$.}
	\label{fig:prompt_token}   
\end{figure}

As stated in previous research study\cite{gao2021making}, the selection of prompts may have a huge impact on the model performance. Therefore, we conduct experiments using different number of prompt tokens on domain $\mathbb{DI}$ $\to$ $\mathbb{AS}$ to further investigate the influence of soft prompts. The results are shown in Figure \ref{fig:prompt_token}, which demonstrate that the length of prompt token effects the performance of prompt tuning on domain $\mathbb{DI}$ $\to$ $\mathbb{AS}$. In our method, we set the prompt token length as 3 to achieve the best results for cross-domain ATE.

\subsection{Case Study}
To further demonstrate the effectiveness of the proposed model, we perform case study on dataset $\mathbb{G}$1. Table \ref{tb:case_study} present the results of cross-domain ATE by AD-SAL, CrossBERT-UDA, and our method. In example 1 ``\textit{Straight-forward, no surprises, very decent Japanese food.}", both CrossBERT-UDA and AD-SAL can not identify the aspect term \textit{Japanese food}. CrossBERT-UDA extracts one of three aspect terms, while AD-SAL only predicts one wrong aspect term in example 2 ``\textit{While there’s a decent menu, it shouldn’t take ten minutes to get your drink and 45 for a dessert pizza}", in which there are three aspect terms \textit{menu}, \textit{drink}, and \textit{dessert pizza}. For example 3 and 6, there are multiple aspect terms and most of them include more than one word. AD-SAL and CrossBERT-UDA can only detect part of words for these aspect terms. In example 4 and 5, baseline models are able to correctly predict part of aspect terms while our method identify all of them. The case study demonstrates that our method can accurately detect not only multiple aspect terms but also aspect terms with multiple words.

\begin{table}
	\centering
	\caption{The prediction results of aspect term on domain $\mathbb{R}$ by AD-SAL, CrossBERT-UDA, and our method. Incorrect predictions are indicated by marker \xmark.}
	\label{tb:case_study}
	\resizebox{\textwidth}{!}{
		\begin{tabular}{l|c|c|c} 
			\hline
			\textbf{Input ($\mathbb{S}$-\textgreater{}$\mathbb{R}$)}                                                                                                                                       & \textbf{AD-SAL}                               & \textbf{CrossBERT-UDA}                                    & \textbf{Ours}                                                                    \\ 
			\hline\hline
			\begin{tabular}[c]{@{}l@{}}Straight-forward, no surprises, very decent \\{[}Japanese food].\end{tabular}                                                                    & \{{[} ]\}                                                                            & \{{[} ]\}                                                                                            & \{{[}Japanese food]\}                                                                                                      \\ 
			\hline
			\begin{tabular}[c]{@{}l@{}}While there’s a decent [menu], it shouldn’t \\take ten minutes to get your [drink] and \\45 for a [dessert pizza].\end{tabular}                 & \{{[}pizza]\xmark\}                                                                        & \{{[}menu]\}                                                                                         & \begin{tabular}[c]{@{}c@{}}\{{[}menu], \\{[}drink], \\{[}dessert pizza]\}\end{tabular}                                    \\ 
			\hline
			\begin{tabular}[c]{@{}l@{}}I’ve had the [jellyfish], [horse mackerel], \\the [blue fin tuna]~and the [sake ikura roll] \\among others, and they were all good.\end{tabular} & \begin{tabular}[c]{@{}c@{}}\{{[}jellyfish], \\{[}horse]\xmark, \\{[}tuna]\xmark\}\end{tabular} & \begin{tabular}[c]{@{}c@{}}\{{[}jellyfish], \\{[}erel]\xmark, \\{[}sake]\xmark, \\{[}ura roll]\xmark\}\end{tabular} & \begin{tabular}[c]{@{}c@{}}\{{[}jellyfish],~\\{[}horse mackerel],~\\{[}blue fin tuna],\\{[}sake ikura roll]\}\end{tabular}  \\ 
			\hline
			\begin{tabular}[c]{@{}l@{}}The [food] is top notch , the [service] is \\attentive , and the [atmosphere] is great.\end{tabular}                                             & \begin{tabular}[c]{@{}c@{}}\{{[}food], \\{[}service]\}\end{tabular}                  & \begin{tabular}[c]{@{}c@{}}\{{[}food], \\{[}service]\}\end{tabular}                                  & \begin{tabular}[c]{@{}c@{}}\{{[}food], \\{[}service], \\{[}atmosphere]\}\end{tabular}                                      \\ 
			\hline
			\begin{tabular}[c]{@{}l@{}}Try the [ribs] , sizzling [beef] and couple \\it with [coconut rice].\end{tabular}                                                               & \begin{tabular}[c]{@{}c@{}}\{{[}ribs], \\{[}beef]\}\end{tabular}                     & \begin{tabular}[c]{@{}c@{}}\{{[}ribs], \\{[}beef], \\{[}rice]\xmark\}\end{tabular}                        & \begin{tabular}[c]{@{}c@{}}\{{[}ribs], \\{[}beef], \\{[}coconut rice]\}\end{tabular}                                       \\ 
			\hline
			\begin{tabular}[c]{@{}l@{}}They have a very good \\{[}chicken with avocado] and good [tuna] \\as well .\end{tabular}                                                        & \begin{tabular}[c]{@{}c@{}}\{{[}chicken]\xmark, \\{[}tuna]\}\end{tabular}                 & \begin{tabular}[c]{@{}c@{}}\{{[}avocado]\xmark, \\{[}tuna]\}\end{tabular}                                 & \begin{tabular}[c]{@{}c@{}}\{{[}chicken with \\avocado], \\{[}tuna]\}\end{tabular}                                         \\
			\hline
		\end{tabular}
	}
\end{table}

\section{Conclusion and Future Work}
\label{sec:cf}
In this paper, we propose a novel soft prompt based joint learning method for cross-domain aspect term extraction. The existing approaches are either machine learning or deep learning based, or hard prompt based methods, which suffer from low-quality domain-invariant features or unstable performance on small-scale target datasets. Different from previous methods, soft prompts are applied to learn in-domain knowledge of different domains to enhance the domain-invariant feature representations. Instead of directly integrating syntax information, a self-supervised learning of syntactic feature is designed to learn the structural correspondence between domains for narrowing the domain gap. Our experiments across two group of datasets spanning a range of domains demonstrate the effectiveness of our approach over the existing models for cross domain aspect term extraction. 

In the future, we plan to expand our model to complete more aspect-based sentiment analysis tasks (e.g., opinion term extraction, aspect-opinion term pair extraction, sentiment triplet extraction, etc.). However, it is a challenging research work to design a unified framework to complete all ABSA sub-tasks in different domains due to the complex relations existed among aspect, opinion, and sentiment polarity. Simultaneously, the performance of proposed method can be further improved on small-scale datasets to enhance the robustness.

\bmhead{Acknowledgments}

The authors would like to acknowledge the financial support from Callaghan Innovation (CSITR1902, 2020), New Zealand, without which this research would not have been possible. We are grateful for their contributions to the advancement of science and technology in New Zealand. The authors would also like to thank CAITO.ai for their invaluable partnership and their contributions to the project.

\bibliographystyle{splncs04}
\bibliography{ref}

\end{document}